\documentclass[10pt, twocolumn, letterpaper]{article}

\usepackage[pagenumbers]{cvpr} 

\def\cvprPaperID{1802}
\def\confName{CVPR}
\def\confYear{2023}

\usepackage{supertabular}
\usepackage{colortbl}
\usepackage[utf8]{inputenc} 
\usepackage[T1]{fontenc}    
\usepackage{microtype}      
\usepackage{xcolor}         
\usepackage{xspace}
\usepackage{amsmath,amssymb,amsfonts,dsfont,pifont,bm,bbm,mathrsfs,mathtools,nicefrac}
\usepackage{algorithm,algpseudocode,listings}
\usepackage{wrapfig}
\usepackage{booktabs,multirow,adjustbox,diagbox,threeparttable}
\definecolor{citeblue}{HTML}{0071bc}
\usepackage[pagebackref=true,breaklinks=true,colorlinks=true,citecolor=citeblue,bookmarks=false]{hyperref}
\usepackage{cleveref}  
\usepackage{bbding}

\crefname{section}{Sec.}{Secs.}
\Crefname{section}{Section}{Sections}
\crefname{appendix}{Appendix}{Appendixes}
\crefname{table}{Tab.}{Tabs.}
\Crefname{table}{Table}{Tables}
\crefname{figure}{Fig.}{Figs.}
\Crefname{figure}{Figure}{Figures}
\crefname{equation}{Eq.}{Eqs.}
\Crefname{equation}{Equation}{Equations}
\crefname{algorithm}{Algorithm}{Algorithms}
\definecolor{codegreen}{rgb}{0,0.6,0}
\definecolor{codegray}{rgb}{0.5,0.5,0.5}
\definecolor{codepurple}{rgb}{0.58,0,0.82}
\definecolor{backcolour}{rgb}{1.0,1.0,1.0}
\lstdefinestyle{mystyle}{
    backgroundcolor=\color{backcolour},
    commentstyle=\color{codegreen},
    keywordstyle=\color{magenta},
    numberstyle=\tiny\color{codegray},
    stringstyle=\color{codepurple},
    basicstyle=\ttfamily\scriptsize,
    breakatwhitespace=false,
    breaklines=true,
    captionpos=b,
    keepspaces=true,
    showspaces=false,
    showstringspaces=false,
    showtabs=false,
    tabsize=2
}
\renewcommand{\paragraph}[1]{\vspace{1.25mm}\noindent\textbf{#1}}

\newcommand{\tocite}[1]{\textcolor{red}{[TO CITE]}}
\newcommand{\method}{HPM\xspace}
\newcommand{\supp}{\textit{Supplementary Material}\xspace}

\definecolor{textpurple}{RGB}{135,89,201}

\definecolor{upcolor}{RGB}{57,182,74}
\newcommand{\up}[1]{\textcolor{upcolor}{$\uparrow$ #1}}
\newcommand{\down}[1]{\textcolor{red}{$\downarrow$ #1}}
\definecolor{deemph}{gray}{0.6}
\newcommand{\gc}[1]{\textcolor{deemph}{#1}}
\newcommand{\pub}[1]{\scriptsize\textcolor{deemph}{[#1]}}
\newlength\savewidth\newcommand\shline{\noalign{\global\savewidth\arrayrulewidth
  \global\arrayrulewidth 1pt}\hline\noalign{\global\arrayrulewidth\savewidth}}
\definecolor{Light}{rgb}{0.99, 0.92, 0.95}

\title{Hard Patches Mining for Masked Image Modeling}

\author{
    Haochen~Wang$^{1,3}$ \hspace{0.6mm}
    Kaiyou~Song$^2$ \hspace{0.6mm}
    Junsong~Fan$^{1,4}$ \hspace{0.6mm}
    Yuxi~Wang$^{1,4}$ \hspace{0.6mm}
    Jin~Xie$^2$ \hspace{0.6mm}
    Zhaoxiang~Zhang$^{1,3,4}$\\[8pt]
    $^1$Center for Research on Intelligent Perception and Computing, \\
    National Laboratory of Pattern Recognition, 
    Institute of Automation, Chinese Academy of Sciences \\
    $^2$Megvii Technology \quad
    $^3$University of Chinese Academy of Sciences \\
    $^4$Centre for Artificial Intelligence and Robotics, \\
    Hong Kong Institute of Science \& Innovation, Chinese Academy of Science \\[3pt]
    \small{\texttt{\{wanghaochen2022, junsong.fan, zhaoxiang.zhang\}@ia.ac.cn}}\\
    \small{\texttt{\{songkaiyou, xiejin\}@megvii.com}}\quad
    \small{\texttt{yuxiwang93@gmail.com}}
}

\begin{document}

\maketitle

\begin{abstract}

Masked image modeling (MIM) has attracted much research attention due to its promising potential for learning scalable visual representations.
In typical approaches, models usually focus on predicting specific contents of masked patches, and their performances are highly related to pre-defined mask strategies.
Intuitively, this procedure can be considered as training a student (the model) on solving given problems (predict masked patches).
%
However, we argue that the model should not only focus on solving given problems, but also \textbf{stand in the shoes of a teacher} to produce a more challenging problem by itself.
To this end, we propose Hard Patches Mining (\method), a brand-new framework for MIM pre-training.
We observe that the reconstruction loss can naturally be the metric of the difficulty of the pre-training task.
Therefore, we introduce an auxiliary loss predictor, predicting patch-wise losses first and deciding where to mask next.
It adopts a relative relationship learning strategy to prevent overfitting to exact reconstruction loss values.
%
%
%
Experiments under various settings demonstrate the effectiveness of \method in constructing masked images.
%
%
%
Furthermore, we empirically find that solely introducing the loss prediction objective leads to powerful representations, verifying the efficacy of the ability to be aware of where is hard to reconstruct.\footnote{Code: \url{https://github.com/Haochen-Wang409/HPM}}

\end{abstract}

\section{Introduction}\label{sec:intro}

Self-supervised learning~\cite{he2020momentum, chen2020simple, grill2020bootstrap, chen2021empirical, chen2021exploring}, with the goal of learning scalable feature representations from large-scale datasets without any annotations, has been a research hotspot in computer vision (CV).
Inspired by masked language modeling (MLM)\cite{devlin2018bert, radford2018improving, radford2019language, brown2020language} in natural language processing (NLP), where the model is urged to predict masked words within a sentence, masked image modeling (MIM), the counterpart in CV, has attracted numerous interests of researchers~\cite{he2022masked, yi2022masked, wei2022masked, xie2022masked, kwon2022masked, pang2022masked, dong2022bootstrapped, bao2021beit}.

\begin{figure}[t]
    \centering
    \includegraphics[width=1\linewidth]{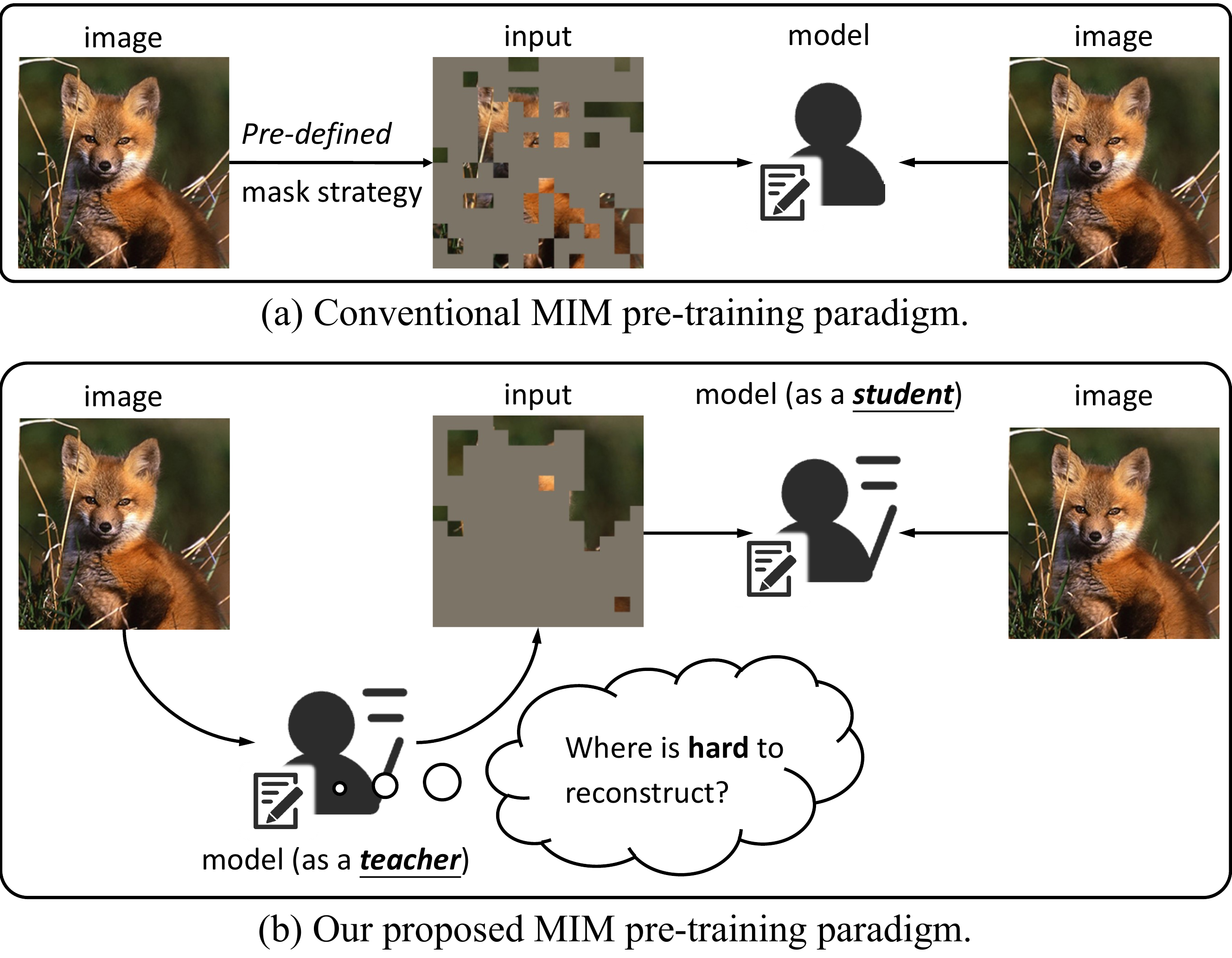}
    \vspace{-20pt}
    \caption{
    Comparison between conventional MIM pre-training paradigm and our proposed \method.
    \textbf{(a)} Conventional approaches can be interpreted as training a \textit{\textbf{student}}, where the model is only equipped with the ability to solve a given problem under some pre-defined mask strategies.
    \textbf{(b)} Our proposed \method pre-training paradigm makes the model to be both a \textbf{\textit{teacher}} and a \textbf{\textit{student}}, with the extra ability to \textit{produce a challenging pretext task}.
    %
    }
    \label{fig:motivation}
    \vspace{-10pt}
\end{figure}

\begin{figure*}[t]
    \centering
    \includegraphics[width=1\linewidth]{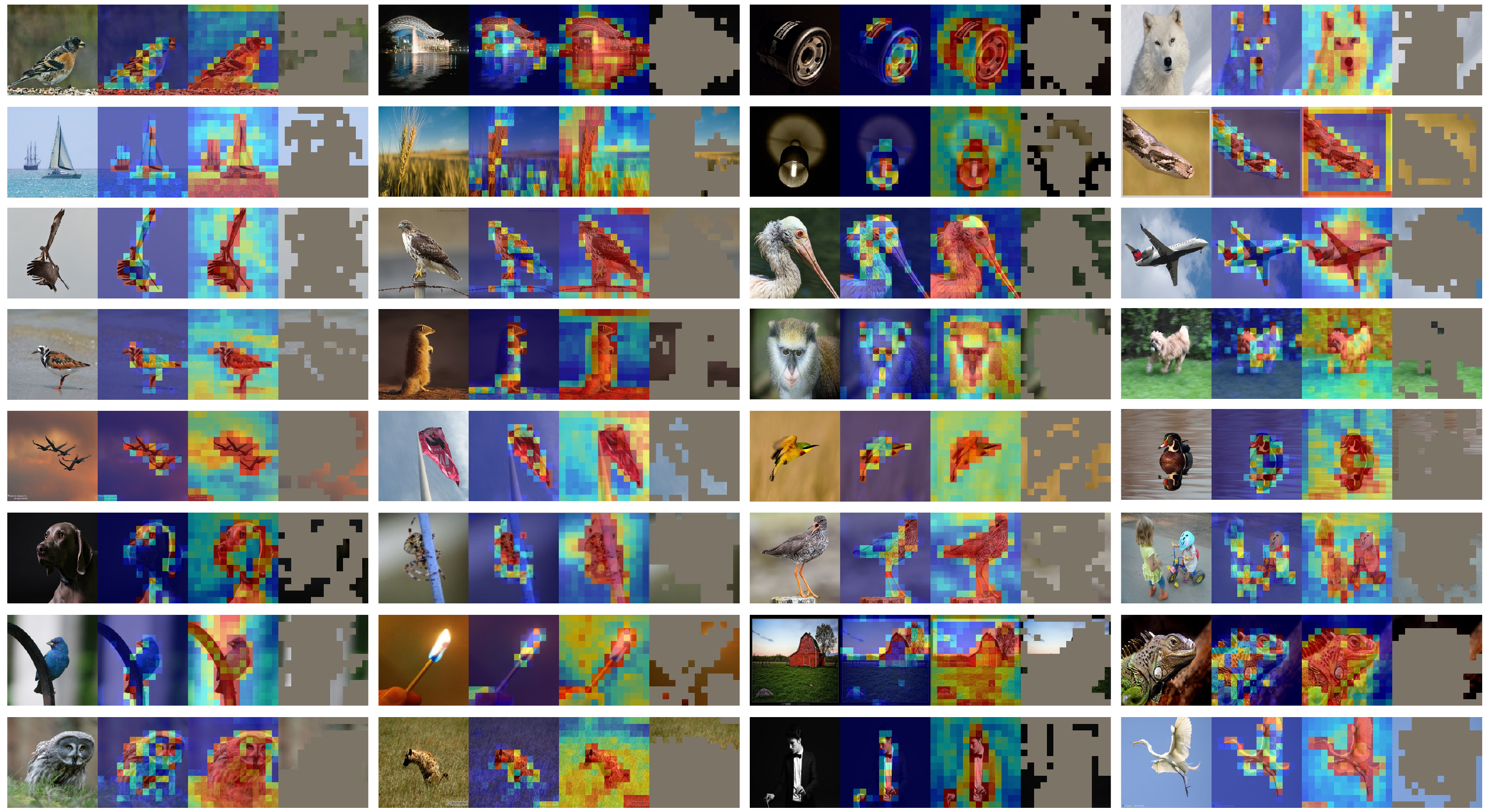}
    \vspace{-18pt}
    \caption{
    Visual comparison between \textbf{reconstruction loss} and \textbf{discriminativeness} on \textbf{ImageNet} \textit{validation} set.
    We load the pre-trained ViT-B/16~\cite{dosovitskiy2020image} provided by MAE~\cite{he2022masked}.
    For each tuple, we show the \textbf{(a)} \textit{input image}, \textbf{(b)} \textit{patch-wise reconstruction loss} averaged over 10 different masks, \textbf{(c)} \textit{predicted loss}, and \textbf{(d)} \textit{masked images} generated by the predicted loss (\textit{i.e.}, patches with top 75\% predicted loss are masked).
    Red means higher loss while blue indicates the opposite.
    \textit{Discriminative parts tend to be hard to reconstruct}.
    }
    \label{fig:discriminative}
    \vspace{-10pt}
\end{figure*}

\cref{fig:motivation}\textcolor{red}{a} illustrates the paradigm of conventional approaches for MIM pre-training~\cite{he2022masked, xie2022simmim, bao2021beit}.
In these typical solutions, models usually focus on predicting specific contents of masked patches.
Intuitively, this procedure can be considered as training a student (\textit{i.e.}, the model) on solving given problems (\textit{i.e.}, predict masked patches).
%
%
To alleviate the spatial redundancy in CV~\cite{he2022masked} and produce a challenging pretext task, mask strategies become critical, which are usually generated under pre-defined manners, \textit{e.g.}, random masking~\cite{he2022masked}, block-wise masking~\cite{bao2021beit}, and uniform masking~\cite{li2022uniform}.
%
%
However, we argue that a difficult pretext task is not all we need, and not only learning to solve the MIM problem is important, but also \textit{learning to produce challenging tasks} is crucial.
In other words, as shown in \cref{fig:motivation}\textcolor{red}{b}, by learning to create challenging problems and solving them \textit{simultaneously}, the model can stand in the shoes of both a \textbf{\textit{student}} and a \textit{\textbf{teacher}}, being forced to hold a more comprehensive understanding of the image contents, and thus leading itself by generating a more desirable task.
%

To this end, we propose \textit{Hard Patches Mining (\method)}, a new training paradigm for MIM.
Specifically, given an input image, instead of generating a binary mask under a manually-designed criterion, we first let the model be a teacher to produce a demanding mask, and then train the model to predict masked patches as a student just like conventional methods.
Through this way, the model is urged to learn where it is worth being masked, and how to solve the problem at the same time.
Then, the question becomes how to design the auxiliary task, to make the model aware of where the hard patches are.

Intuitively, we observe that the reconstruction loss can be naturally a measure of the difficulty of the MIM task, which can be verified by the first two elements of each tuple in \cref{fig:discriminative}, where
the backbone\footnote{\url{https://dl.fbaipublicfiles.com/mae/visualize/mae\_visualize\_vit\_base.pth}} pre-trained by MAE~\cite{he2022masked} with 1600 epochs is used for visualization.
As expected, we find that those discriminative parts of an image (\textit{e.g.}, object) are usually hard to reconstruct, resulting in larger losses.
Therefore, by simply urging the model to  \textit{predict reconstruction loss} for each patch, and then masking those patches with higher predicted losses, we can obtain a more formidable MIM task.
To achieve this, we introduce an auxiliary loss predictor, predicting patch-wise losses first and deciding where to mask next based on its outputs.
To prevent it from being overwhelmed by the exact values of reconstruction losses and make it concentrate on the \textit{relative relationship among patches}, we design a novel relative loss based on binary cross-entropy as the objective.
We further evaluate the effectiveness of the loss predictor using a ViT-B under 200 epochs pre-training in \cref{fig:discriminative}.
As the last two elements for each tuple in \cref{fig:discriminative} suggest, patches with larger \textit{predicted} losses tend to be discriminative, and thus masking these patches brings a challenging situation, where objects are almost masked.
%
%
Meanwhile, considering the training evolution, we come up with an easy-to-hard mask generation strategy, providing some reasonable hints at the early stages. 

Empirically, we observe significant and consistent improvements over the supervised baseline and vanilla MIM pre-training under various settings.
Concretely, with only 800 epochs pre-training, \method achieves 84.2\% and 85.8\% Top-1 accuracy on ImageNet-1K~\cite{russakovsky2015imagenet} using ViT-B and ViT-L, outperforming MAE~\cite{he2022masked} pre-trained with 1600 epochs by +0.6\% and +0.7\%, respectively.
%
\section{Related Work}\label{sec:related}

\noindent\textbf{Self-supervised learning.}
Aiming at learning from data without any annotations, self-supervised learning (SSL) approaches have raised significant interest in computer vision, and how to design an appropriate pretext task becomes the crux~\cite{doersch2015unsupervised, wang2015unsupervised, oord2018representation, zhang2016colorful}.
Among them, contrastive learning~\cite{he2020momentum, oord2018representation, grill2020bootstrap, wang2022semi} based on instance discrimination~\cite{wu2018unsupervised} becomes popular.
The core idea lies in urging the model to learn view-invariant features, and thus these methods strongly depend on data augmentations~\cite{chen2020simple, grill2020bootstrap}.
%
%
MIM pursues a conceptually different direction with different behaviors.

\paragraph{Masked image modeling.}
Since MLM~\cite{devlin2018bert, radford2018improving, radford2019language, brown2020language} and its autoregressive variants have achieved great success in NLP, MIM, its counterpart in CV, has attracted numerous interests of many researchers~\cite{bao2021beit, he2022masked, wei2022masked, zhou2021ibot, xie2022simmim, chen2022context}, with the goal of building a unified self-supervised pre-training framework.
Specifically, for MIM, a Vision Transformer (\textit{e.g.}, ViT~\cite{dosovitskiy2020image} or its hierarchical variants~\cite{liu2021swin, wang2021pyramid, liu2022swin}) is trained to predict pre-defined targets (\textit{e.g.}, discrete tokens~\cite{bao2021beit} generated by a dVAE~\cite{rolfe2016discrete} pre-trained on DALLE~\cite{ramesh2021zero}, raw RGB pixels~\cite{he2022masked, xie2022simmim, liu2022mixmim, li2022uniform}, HoG features~\cite{wei2022masked}, frequency~\cite{liu2022devil, xie2022masked}, and features from a momentum teacher~\cite{zhou2021ibot, baevski2022data2vec, yi2022masked, wu2022extreme, dong2022bootstrapped}) of masked patches.
Also, it has been verified to be an efficient pre-training framework in video understanding~\cite{wang2022bevt, tong2022videomae}, cross-modality~\cite{hou2022milan, bachmann2022multimae, geng2022multimodal, kwon2022masked}, and 3D cases~\cite{yu2022point, pang2022masked, liu2022masked, min2022voxel}.

\paragraph{Mask strategies in masked image modeling.}
In NLP, a word is already highly semantic, and thus vanilla random masking brings a challenging pretext task~\cite{devlin2018bert, dosovitskiy2020image},
By contrast, the success of masked image modeling heavily relies on the mask strategies due to the spatial information redundancy~\cite{he2022masked} in computer vision.
Concretely, MAE~\cite{he2022masked} uses a large mask ratio (\textit{i.e.}, 75\%), BEiT~\cite{bao2021beit} adopts block-wise masking, and SimMIM~\cite{xie2022simmim} finds that larger mask kernels (\textit{e.g.}, 32$\times$32) are more robust against different mask ratios.
Furthermore, AttMask~\cite{kakogeorgiou2022hide} masks patches with high attention signals, bringing a more challenging pretext task.
ADIOS~\cite{shi2022adversarial} trains an extra U-Net~\cite{ronneberger2015u} based masking model by adversarial objectives.
SemMAE~\cite{li2022semmae} regards semantic parts as the visual analog of words, and trains an extra StyleGAN~\cite{karras2019style} based decoder distilled by iBOT~\cite{zhou2021ibot}.
UM-MAE~\cite{li2022uniform} masks one patch in each 2$\times$2 local window, enabling pyramid-based ViTs (\textit{e.g.}, PVT~\cite{wang2021pyramid}, CoaT~\cite{xu2021co}, and Swin~\cite{liu2021swin, liu2022swin}) to take the random sequence of partial vision tokens as input.
All the masking models of these methods are either pre-defined~\cite{he2022masked, wei2022masked, xie2022simmim, bao2021beit, zhou2021ibot, baevski2022data2vec, kakogeorgiou2022hide} or separately learned~\cite{shi2022adversarial, li2022semmae}.
However, we argue that \textit{learn to mask} the discriminative parts is crucial, which can not only guide the model in a more challenging manner, but also bring salient prior of input images, bootstrapping the performance on a wide range of downstream tasks hence.
\section{Method}\label{sec:method}

\begin{figure*}[t]
    \centering
    \includegraphics[width=0.9\linewidth]{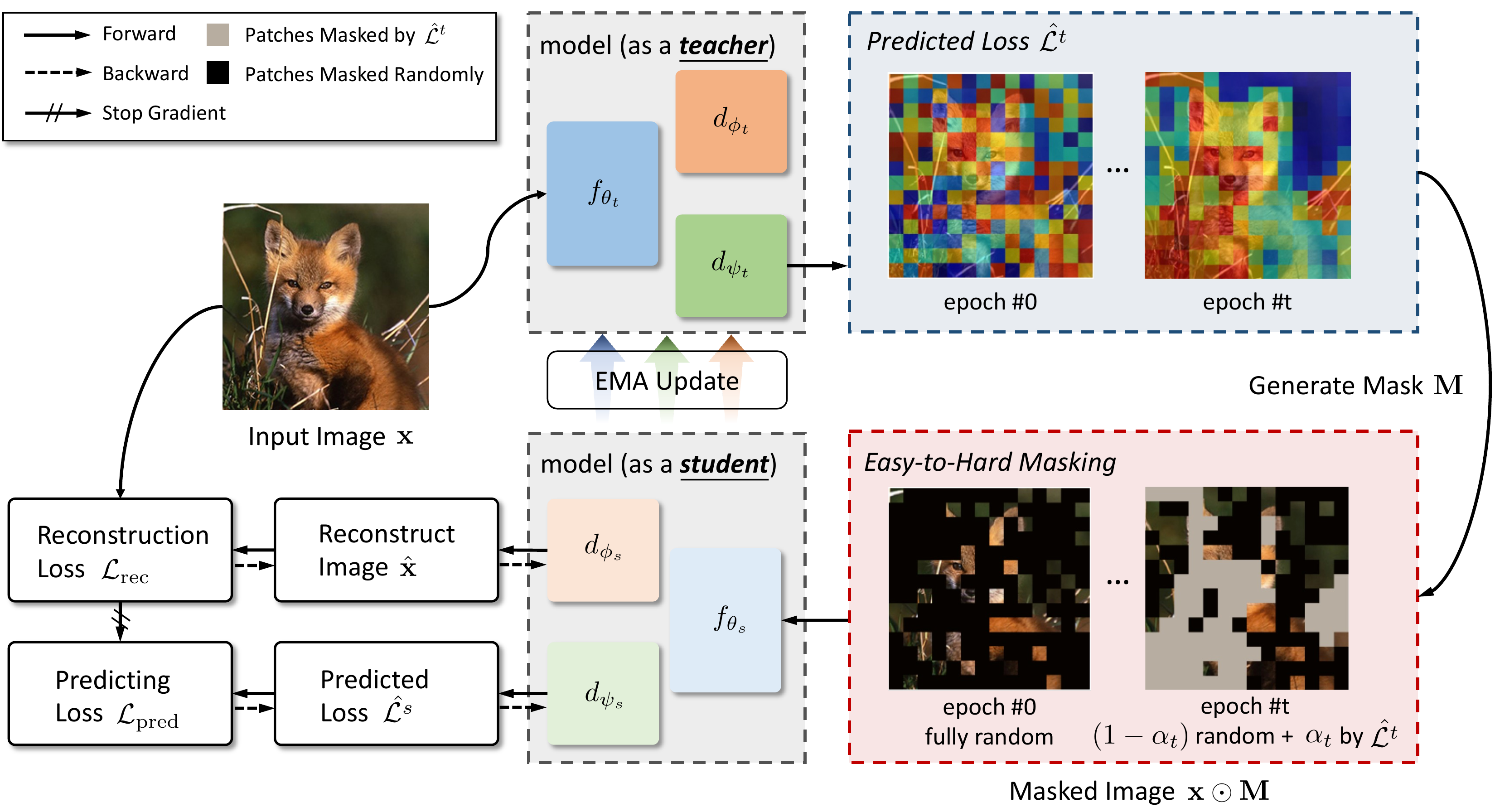}
    \vspace{-5pt}
    \caption{
    \textbf{Illustration of our proposed \method},
    containing a student network and a teacher network, where the teacher is updated by the student in an exponential moving average (EMA) manner.
    Each network consists of an encoder $f_{\theta}$, an image reconstructer $d_{\phi}$, and a loss predictor $d_{\psi}$, parameterized by $\theta$, $\phi$, and $\psi$, respectively.
    For each image during pre-training, it is first fed into the teacher to predict the patch-wise reconstruction loss.
    Then, a binary mask is generated based on the current epoch and the predicted loss.
    Finally, only visible patches are fed into the student to 1) reconstruct masked patches defined in \cref{eq:mim}, and 2) predict relative loss defined in \cref{eq:bce}.
    %
    }
    \label{fig:pipeline}
\end{figure*}

In this section, we first give an overview of our proposed \method in \cref{sec:overview}.
Then, the two objectives in \method, \textit{i.e.}, reconstruction loss and predicting loss are introduced in \cref{sec:mim} and \cref{sec:loss_pred}, respectively.
Finally, in \cref{sec:mask}, the easy-to-hard mask generation manner is described, together with the pseudo-code of the overall training procedure.

\subsection{Overview}
\label{sec:overview}

Introduced in \cref{fig:motivation} and \cref{sec:intro}, conventional MIM pre-training solutions can be considered as training a student to solve \textit{given} problems,
while we argue that \textit{making the model stand in the shoes of a teacher}, producing challenging pretext task is crucial.
To achieve this, we introduce an auxiliary decoder to predict the reconstruction loss of each masked patch, and carefully design its objective.
\cref{fig:pipeline} gives an overview of our proposed \method, introduced next.

\method consists of a student ($f_{\theta_s}$, $d_{\phi_s}$, and $d_{\psi_s}$) and a teacher ($f_{\theta_t}$, $d_{\phi_t}$, and $d_{\psi_t}$) with the same network architecture.
$f_{\theta}(\cdot)$, $d_{\phi}(\cdot)$, and $d_{\psi}(\cdot)$ are encoder, image reconstructor, and reconstruction loss predictor, parameterized by $\theta$, $\phi$, and $\psi$, respectively.
The subscript $t$ stands for teacher and $s$ stands for student.
To generate consistent predictions (especially for the reconstruction loss predictor), momentum update~\cite{he2020momentum} is applied to the teacher:
\begin{equation}
    \bm{\theta}_t \leftarrow m\bm{\theta}_t + (1 - m)\bm{\theta}_s,
\end{equation}
where $\bm{\theta}_t = (\theta_t, \phi_t, \psi_t)$, $\bm{\theta}_s = (\theta_s, \phi_s, \psi_s)$, and $m$ denotes the momentum coefficient.

At each training iteration, an input image $\mathbf{I} \in \mathbb{R}^{H\times W \times C}$ is reshaped into a sequence of 2D patches $\mathbf{x} \in \mathbb{R}^{N\times(P^2C)}$.
$(H,W)$ is the resolution of the original image, $C$ is the number of channels, $P$ is the patch size (\textit{e.g.}, 16), and $N=HW/P^2$ hence.
Then, $\mathbf{x}$ is fed into the teacher to get patch-wise predicted reconstruction loss $\hat{\mathcal{L}}^t = d_{\psi_t}(f_{\theta_t}(\mathbf{x}))$ described in \cref{sec:mim}.
Based on predicted reconstruction loss $\hat{\mathcal{L}}^t$ and the training status, a binary mask $\mathbf{M} \in \{0,1\}^{N}$ is generated under an easy-to-hard manner introduced later in \cref{sec:mask}.
The student is trained based on two objectives, \textit{i.e.}, reconstruction loss (\cref{sec:mim}) and predicting loss (\cref{sec:loss_pred})
\begin{equation}
    \mathcal{L} = \mathcal{L}_{\mathrm{rec}} + \mathcal{L}_{\mathrm{pred}},
\end{equation}
where these two objectives work in an alternating way, and reinforce each other to extract better representations, by gradually urging the student to reconstruct hard patches within an image.

\subsection{Image Reconstructor}
\label{sec:mim}
Masked image modeling aims at training an autoencoder (\textit{i.e.}, image reconstructor) to reconstruct the masked portion according to pre-defined targets, \textit{e.g.}, raw RGB pixels~\cite{he2022masked, xie2022simmim, liu2022exploring, kong2022understanding, chen2022context, yi2022masked} and specific features~\cite{bao2021beit, zhou2021ibot, wei2022masked, baevski2022data2vec, wu2022extreme}.
\begin{equation}
\label{eq:mim}
    \mathcal{L}_{\mathrm{rec}} = \mathcal{M} \left(
    d_{\phi_s}(f_{\theta_s}(\mathbf{x} \odot \mathbf{M})), \mathcal{T}(\mathbf{x} \odot (1 - \mathbf{M})) 
    \right),
\end{equation}
where for conventional approaches, the binary mask $\mathbf{M} \in \{0,1\}^{N}$ is generated by a pre-defined manner.
$\odot$ means element-wise dot product, and thus $\mathbf{x} \odot \mathbf{M}$ represents unmasked (\textit{i.e.}, visible) patches and vice versa.
$\mathcal{T}(\cdot)$ is the transformation function, generating reconstructed targets.
$\mathcal{M}(\cdot,\cdot)$ represents the similarity measurement, \textit{e.g.}, $\ell_2$-distance~\cite{he2022masked}, smooth $\ell_1$-distance~\cite{xie2022simmim}, knowledge distillation~\cite{zhou2021ibot, dong2022bootstrapped}, and cross-entropy~\cite{bao2021beit}.

\subsection{Hard Patches Mining with a Loss Predictor}
\label{sec:loss_pred}
It is widely known that in NLP, each word in a sentence is already highly semantic~\cite{he2022masked}.
Training a model to predict only a few missing words tends to be a challenging task in understanding languages~\cite{devlin2018bert, brown2020language, radford2018improving, radford2019language}.
While in CV, on the contrary, an image is with heavy spatial redundancy, and thus plenty of mask strategies are proposed to deal with this issue~\cite{he2022masked, xie2022simmim, bao2021beit, kakogeorgiou2022hide, shi2022adversarial, li2022semmae}.

Apart from designing a challenging situation by prior knowledge, we argue that \textit{the ability to produce demanding scenarios} is also crucial for MIM pre-training.
Intuitively, we consider patches with high reconstruction loss defined in \cref{eq:mim} as \textit{hard patches}, which implicitly indicate the most discriminative parts of an image, which is verified in \cref{fig:discriminative}.
%
%
Therefore, if the model is equipped with the ability to predict the reconstruction loss for each patch, simply masking those hard patches becomes a more challenging pretext task.

To this end, we employ an extra loss predictor (\textit{i.e.}, $d_{\psi}$ in \cref{fig:pipeline}) to mine hard patches during training.
Next, we will introduce how to design the objective for loss predictor with two variants: 1) absolute loss and 2) relative loss.

\paragraph{Absolute loss.}
The simplest and the most straightforward way is to define the objective in an MSE manner.
\begin{equation}
\label{eq:mse}
    \mathcal{L}_{\mathrm{pred}} = \left(
    d_{\psi_s}(f_{\theta_s} (\mathbf{x} \odot \mathbf{M})) - \mathcal{L}_{\mathrm{rec}}
    \right)^2 \odot (1 - \mathbf{M}),
\end{equation}
where $d_{\psi_s}$ is the auxiliary decoder of the student parameterized by $\psi_s$, and $\mathcal{L}_{\mathrm{rec}}$ here is detached from gradient, being a ground-truth for loss prediction.
However, recall that our goal is to determine \textit{hard patches} within an image, thus we need to learn the \textit{relative relationship among patches}.
Under such a setting, MSE is not the most suitable choice hence, since the scale of $\mathcal{L}_{\mathrm{rec}}$ decreases as training goes on, and thus the loss predictor may be overwhelmed by the scale and the exact value of $\mathcal{L}_{\mathrm{rec}}$.
For this purpose, we propose a binary cross-entropy-based relative loss as an alternative.

\paragraph{Relative loss.}
Given a sequence of reconstruction loss $\mathcal{L}_{\mathrm{rec}}\in\mathbb{R}^N$, we aim to predict $\texttt{argsort}(\mathcal{L}_{\mathrm{rec}})$ by using a relative loss.
That is because, within an image, the patch-wise difficulty of the reconstruction task can be measured by $\texttt{argsort} (\mathcal{L}_{\mathrm{rec}})$.
However, as the \texttt{argsort}$(\cdot)$ operation is non-differentiable, it is hard to directly minimize some custom distances between $\texttt{argsort}(d_{\psi_s}(f_{\theta_s} (\mathbf{x} \odot \mathbf{M})))$ and $\texttt{argsort}(\mathcal{L}_{\mathrm{rec}})$.
Therefore, we translate this problem into an equivalent one: \textit{dense relation comparison}.
Specifically, for each pair of patches $(i,j)$, where $i,j=1,2,\cdots,N$ and $i \neq j$, we can implicitly learn $\texttt{argsort}(\mathcal{L}_{\mathrm{rec}})$ by predicting the relative relation of $\mathcal{L}_{\mathrm{rec}}(i)$ and $\mathcal{L}_{\mathrm{rec}}(j)$, \textit{i.e.}, which one is larger.
The objective is defined as follows:
\begin{equation}
\label{eq:bce}
\begin{aligned}
    \mathcal{L}_{\mathrm{pred}} = 
    &-\sum_{i=1}^N \sum_{j=1 \atop j\neq i}^N \mathbbm{1}^{+}_{ij} \log \left( \sigma(\hat{\mathcal{L}}^s_i - \hat{\mathcal{L}}^s_j) \right) \\
    &-\sum_{i=1}^N \sum_{j=1 \atop j\neq i}^N \mathbbm{1}^{-}_{ij} \log \left( 1 - \sigma(\hat{\mathcal{L}}^s_i - \hat{\mathcal{L}}^s_j) \right),
\end{aligned}
\end{equation}
where $\hat{\mathcal{L}}^s = d_{\psi_s}(f_{\theta_s}(\mathbf{x} \odot \mathbf{M})) \in \mathbb{R}^N$ represents the predicted loss from the student, and $i, j=1,2,\dots,N$ are patch indexes.
$\sigma(\cdot)$ indicates \texttt{sigmoid} function, \textit{i.e.}, $\sigma(z) = e^z / (e^z + 1)$.
$\mathbbm{1}^{+}_{ij}$ and $\mathbbm{1}^{-}_{ij}$ are two indicators, representing the relative relationship of ground-truth reconstruction losses, \textit{i.e.}, $\mathcal{L}_{\mathrm{rec}}$, between patch $i$ and patch $j$
\begin{equation}
\label{eq:indicatorp}
\mathbbm{1}^{+}_{ij} = \left\{
\begin{aligned}
    &1, &&\mathcal{L}_{\mathrm{rec}}(i) > \mathcal{L}_{\mathrm{rec}}(j) \mathrm{\ and\ } \mathbf{M}_i=\mathbf{M}_j=0, \\
    &0, &&\mathrm{otherwise},
\end{aligned}
\right.
\end{equation}
\begin{equation}
\label{eq:indicatorn}
\mathbbm{1}^{-}_{ij} = \left\{
\begin{aligned}
    &1, &&\mathcal{L}_{\mathrm{rec}}(i) < \mathcal{L}_{\mathrm{rec}}(j) \mathrm{\ and\ } \mathbf{M}_i=\mathbf{M}_j=0, \\
    &0, &&\mathrm{otherwise},
\end{aligned}
\right.
\end{equation}
where $\mathbf{M}_i=\mathbf{M}_j=0$ means that both patch $i$ and $j$ are masked during training.

\subsection{Easy-to-Hard Mask Generation}
\label{sec:mask}
With the reconstruction loss predictor, we are able to define a more challenging pretext task, \textit{i.e.}, mask those hard/discriminative parts of an input image.
Concretely, after obtaining the predicted reconstruction loss from the teacher network, \textit{i.e.}, $\hat{\mathcal{L}}^t = d_{\psi_t}(f_{\theta_t} (\mathbf{x}))$, we conduct $\texttt{argsort}(\cdot)$ operation over $\hat{\mathcal{L}}^t$ in a descending order to obtain the relative reconstruction difficulty within the image.

However, in the early training stages, the learned feature representations are not ready for reconstruction but are overwhelmed by the rich texture, which means large reconstruction loss may not be equivalent to discriminative.
To this end, we propose an easy-to-hard mask generation manner, providing some reasonable hints that guide the model to reconstruct masked hard patches step by step.

As illustrated in \cref{fig:pipeline}, for each training epoch $t$, $\alpha_t$ of the mask patches are generated by $\hat{\mathcal{L}}^t$, and the remaining $1 - \alpha_t$ are randomly selected.
Specifically, $\alpha_t = \alpha_0 + \frac{t}{T} (\alpha_T - \alpha_0)$,
where $T$ is the total training epochs, and $\alpha_0,\alpha_T\in[0,1]$ are two tunable hyper-parameters.
We filter $\alpha_t\cdot\gamma N$ patches with the highest $\hat{\mathcal{L}}^t$ to be masked, and the remaining $(1-\alpha_t)\cdot\gamma N$ patches are randomly masked.
The proportion $\alpha_t$ gradually increases from $\alpha_0$ to $\alpha_T$ in a linear manner without further tuning for simplicity, contributing to an easy-to-hard training procedure.

\cref{alg:code} summarizes the training procedure, together with the pseudo-code of computing the objective for training the reconstruction loss predictor.
Thanks to the simple implementation of the easy-to-hard mask generation, please refer to \supp for the pseudo-code.

\begin{algorithm}[t]
\caption{Pseudo-Code of \method in a PyTorch-like Style.}
\label{alg:code}
\begin{lstlisting}[language=python]
# model_s, model_t: networks for student and teacher
# t, T: current/total epochs
# x: input patchified images
# rec: reconstructed image
# pred: predicted reconstruction loss

# teacher inference
_, pred_t = model_t(x)
# easy-to-hard mask generation
mask = mask_generation(pred_t, t, T, mask_ratio)
# student forward to compute objectives
rec_x, pred_s = model_s(x * mask)
# compute losses
loss_rec = (rec_x - x[~mask]) ** 2
loss_pred = predicting_loss(pred_s, loss_rec, mask)
return loss_rec + loss_pred

# predict relative reconstruction loss
def predicting_loss(pred_s, loss_rec, mask):
    loss_rec = loss_rec[~mask].detach()
    pred_s = pred_s[~mask]
    # generate indicators
    pos = loss_rec.unsqueeze(0) > loss_rec.unsqueeze(1)
    neg = loss_rec.unsqueeze(0) < loss_rec.unsqueeze(1)
    valid = pos + neg
    # compute dense relative relationship
    pred_mat = pred_s.unsqueeze(0) > pred_s.unsqueeze(1)
    # compute predicting loss
    loss_pos = -pos * log(sigmoid(pred_mat))
    loss_neg = -neg * log(1-sigmoid(pred_mat))
    loss = loss_pos + loss_neg
    return loss.sum() / valid.sum()
\end{lstlisting}
\end{algorithm}


\section{Experiments}\label{sec:exp}


\noindent\textbf{Baseline.}
We evaluate our proposed \method under self-supervised pre-training on ImageNet-1K~\cite{russakovsky2015imagenet}.
We take ViT-B/16~\cite{dosovitskiy2020image} as the backbone and MAE~\cite{he2022masked} pre-trained with 200 epochs on ImageNet-1K~\cite{russakovsky2015imagenet} as our baseline.
Our implementation is based on MAE~\cite{he2022masked} and UM-MAE~\cite{li2022uniform}.
More details can be found in \supp.

\paragraph{ImageNet classification.}
We evaluate our proposed \method by 1) end-to-end fine-tuning, 2) linear probing, and 3) $k$-NN.
We report Top-1 accuracy (\%) on the validation set.
End-to-end fine-tuning (or learning from scratch) and linear probing over image classification are trained for 100 epochs.
$k$-NN is implemented based on DINO~\cite{caron2021emerging}.
The resolution is kept to 224$\times$224 on both pre-training and evaluation.

\paragraph{COCO object detection and instance segmentation.}
We take Mask R-CNN~\cite{he2017mask} with FPN~\cite{lin2017feature} as the object detector, and perform end-to-end fine-tuning on COCO~\cite{lin2014microsoft} for 1$\times$ schedule (12 epochs) for ablations (\textit{i.e.}, \cref{tab:downstream}) with 1024$\times$1024 resolution.
We report AP$_{\text{box}}$ for object detection and AP$_{\text{mask}}$ for instance segmentation.
Our implementation is based on detectron2~\cite{wu2019detectron2} and ViTDet~\cite{li2021benchmarking}.

\paragraph{ADE20k semantic segmentation.}
We take UperNet~\cite{xiao2018unified} as the segmentor, and perform end-to-end fine-tuning on ADE20k~\cite{zhou2017scene} for 80k iterations for ablations (\textit{i.e.}, \cref{tab:downstream}) and 160k iterations when comparing with previous methods (\textit{i.e.}, \cref{tab:sota_seg}) with 512$\times$512 resolution.
We take mIoU~\cite{everingham2015pascal} as the evaluation metric.
Our implementation is based on mmsegmentation~\cite{mmseg2020}.

\subsection{Ablation Study}
\label{sec:ablation}

We study different reconstruction targets, mask strategies, predicting loss formulations, and downstream tasks in this section.
By default, ViT-B/16~\cite{dosovitskiy2020image} is used as the backbone with 200 epochs pre-training and 100 epochs fine-tuning on ImageNet-1K~\cite{russakovsky2015imagenet}.
We \colorbox{Light}{highlight} our default settings.
%

\paragraph{Reconstruction targets.}
We study the effectiveness of different reconstruction targets in \cref{tab:target}, including regressing raw RGB pixels used in MAE~\cite{he2022masked}, and distilling from various teacher models, \textit{i.e.}, the EMA (exponential moving average) teacher used in BootMAE~\cite{dong2022bootstrapped}, and pre-trained teachers obtained from DINO~\cite{caron2021emerging} and CLIP~\cite{radford2021learning}.
All these teacher models share the same architecture, \textit{i.e.}, ViT-B/16~\cite{dosovitskiy2020image}.


It has been substantiated that directly regressing RGB values of pixels is a simple yet efficient way in MIM pre-training~\cite{he2022masked}.
However, due to the existence of high-frequency noise in some cases, patches with higher frequency tend to have larger reconstruction loss, and thus \textit{hard patches may not be highly semantic} under this setting, which is quite the opposite from our motivation: learn to mine \textit{discriminative} parts of an image instead of high-frequency parts.
To this end, we further take features from a teacher model to be the learning target (\textit{e.g.}, DINO~\cite{caron2021emerging} and CLIP~\cite{radford2021learning}), to verify the effectiveness of our proposed \method.
%

Note that the objective differs when using different reconstruction targets.
Specifically, an MSE loss is adopted for RGB regression following MAE~\cite{he2022masked}, while for knowledge distillation cases, we first apply $\ell_2$ normalization to the features output from the teacher and the student, and then minimize their MSE distances.
This can be also implemented by maximizing their cosine similarities.

\begin{table}[t]
    \centering
    \caption{
    Ablation study on different \textbf{reconstruction targets}.
    We study four different targets, including raw RGB pixels (MAE~\cite{he2022masked} baseline), and three knowledge distillation targets, \textit{i.e.}, features from the EMA (exponential moving average) model, DINO~\cite{caron2021emerging}, and CLIP~\cite{radford2021learning}.
    All cases are pre-trained 200 epochs on ImageNet~\cite{russakovsky2015imagenet} with ViT-B/16~\cite{dosovitskiy2020image}.
    %
    %
    %
    }
    \label{tab:target}
    \vspace{-8pt}
    \setlength{\tabcolsep}{4pt}
    \scalebox{0.85}{
    \begin{tabular}{l|ll|lll}
    \toprule
    \multirow{2}{*}{target} & \multirow{2}{*}{$\mathcal{L}_{\mathrm{pred}}$} & learn & \multirow{2}{*}{fine-tune} & \multirow{2}{*}{\gc{linear}} & \multirow{2}{*}{\gc{$k$-NN}} \\
    & & to mask & & & \\
    \midrule
    \multicolumn{6}{l}{\textit{Pixel Regression}} \\
    \midrule
    \multirow{3}{*}{RGB (MAE~\cite{he2022masked}} & - & - & 82.23 & \gc{50.80} & \gc{29.84} \\
    & \checkmark & - & 82.49 \up{0.26} & \gc{51.26} & \gc{31.98} \\
    & \cellcolor{Light}{\checkmark} & \cellcolor{Light}{\checkmark} & \cellcolor{Light}{\textbf{82.95 \up{0.72}}} & \cellcolor{Light}{\gc{\textbf{54.92}}} & \cellcolor{Light}{\gc{\textbf{36.09}}} \\
    \midrule
    \multicolumn{6}{l}{\textit{Feature Distillation}} \\
    \midrule
    \multirow{3}{*}{EMA features} & - & - & 82.99 & \gc{32.65} & \gc{20.69} \\
    & \checkmark & - & 83.13 \up{0.14} & \gc{52.06} & \gc{35.73} \\
    & \checkmark & \checkmark & \textbf{83.47 \up{0.48}} & \gc{\textbf{55.25}} & \gc{\textbf{35.94}} \\
    \midrule
    \multirow{3}{*}{DINO~\cite{caron2021emerging} features} & - & - & 83.46 & \gc{61.31} & \gc{41.53} \\
    & \checkmark & - & 83.58 \up{0.12} & \gc{63.25} & \gc{43.02} \\
    & \checkmark & \checkmark & \textbf{84.13 \up{0.67}} & \gc{\textbf{64.17}} & \gc{\textbf{47.25}} \\
    \midrule
    \multirow{3}{*}{CLIP~\cite{radford2021learning} features} & - & - & 83.20 & \gc{59.80} & \gc{42.51} \\
    & \checkmark & - & 83.31 \up{0.11} & \gc{60.62} & \gc{43.26} \\
    & \checkmark & \checkmark & \textbf{83.58 \up{0.38}} & \gc{\textbf{62.22}} & \gc{\textbf{45.08}} \\
    \bottomrule
    \end{tabular}}
\end{table}

\begin{table}[t]
    \centering
    \caption{
    Ablation study on different \textbf{mask strategies}.
    We study the effect of different $\alpha_0$, $\alpha_T$, and $\gamma$.
    Large $\alpha_T$ indicates a more difficult pretext task, but the randomness of this strategy decreases.
    %
    }
    \label{tab:mask}
    \vspace{-8pt}
    \setlength{\tabcolsep}{3.1pt}
    \scalebox{0.85}{
    \begin{tabular}{l|cc|lll|l}
    \toprule
    case & difficulty & randomness & $\gamma$ & $\alpha_0$ & $\alpha_T$ & fine-tune \\
    \midrule
    random & easy & strong & 75 & 0 & 0 & 82.49 \\
    learn to mask & \multirow{2}{*}{$\Big\downarrow$} & \multirow{2}{*}{$\Big\downarrow$} & \cellcolor{Light}{75} & \cellcolor{Light}{0} & \cellcolor{Light}{0.5} & \cellcolor{Light}{\textbf{82.95 \up{0.46}}} \\
    learn to mask & & & 75 & 0 & 1 & 82.67 \up{0.18} \\
    learn to mask & hard & weak & 75 & 1 & 1 & 81.40 \down{1.09} \\
    \midrule
    random & easy & strong & 50 & 0 & 0 & 82.36 \\
    learn to mask & $\downarrow$ & $\downarrow$ & 50 & 0 & 0.5 & \textbf{82.56 \up{0.20}} \\
    learn to mask & hard & weak & 50 & 1 & 1 & 82.19 \down{0.17} \\
    \midrule
    random & easy & strong & 90 & 0 & 0 & 82.48 \\
    learn to mask & $\downarrow$ & $\downarrow$ & 90 & 0 & 0.5 & \textbf{82.66 \up{0.18}} \\
    learn to mask & hard & weak & 90 & 1 & 1 & 80.59 \down{1.89} \\
    \bottomrule
    \end{tabular}}
    \vspace{-10pt}
\end{table}

As illustrated in \cref{tab:target}, our \method is able to bootstrap the performances under various learning targets.
Taking the pixel regression case as an instance, equipped with the predicting loss and the easy-to-hard mask generation manner, the fine-tuning Top-1 accuracy achieves 82.95\%, outperforming MAE~\cite{he2022masked} by +0.72\%.
Notably, \textit{only} applying an auxiliary decoder to predict reconstruction loss for each patch brings an improvement of +0.26\% fine-tuning accuracy, achieving 82.49\%, verifying that \textit{the ability to mine hard patches} brings better extracted feature representations.
Then, fully taking advantage of this capability, \textit{i.e.}, generate challenging masks, can further bootstrap the performances,
which appears \textit{consistently across different learning targets}.

\paragraph{Mask strategies.}
To verify that harder tasks do bring better performance, we study various mask strategies in \cref{tab:mask}, including random masking and our proposed learnable masking.
With different $\alpha_0$ and $\alpha_T$, we can construct different strategies.
For instance, $\alpha_0=\alpha_T=0$ indicates that predicted reconstruction losses $\hat{\mathcal{L}}^t$ will not participate in mask generation (\textit{i.e.}, a fully random manner),
$\alpha_0=\alpha_T=1$, however, means that $\gamma N$ patches with the highest $\hat{\mathcal{L}}^t$ values are kept masked (see \cref{fig:discriminative}).

From \cref{tab:mask}, we find that the increase in the difficulty of the pretext task does not consistently lead to better performance.
\textit{Retaining a certain degree of randomness} is beneficial for satisfactory results.
Specifically, $\alpha_0=0$ and $\alpha_T=0.5$ achieves the best results under different mask ratio $\gamma$, which is a more difficult case over $\alpha=\alpha_T=0$ (\textit{i.e.}, random masking), and with stronger randomness against $\alpha_0=\alpha_T=1$.
These conclusions are quite intuitive.
Directly masking those patches with the highest $\hat{\mathcal{L}}^t$ brings the hardest problem, where discriminative parts of an image are almost masked.
That means visible patches are nearly all background (see \cref{fig:discriminative}).
\textit{Forcing the model to reconstruct the forehead based on only these backgrounds without any hints makes no sense},
whose performance drops consistently with different values of $\gamma$.
Therefore, a certain level of randomness is necessary.
%

We further investigate the effectiveness of producing \textit{hard} pretext task for MIM pre-training in \cref{tab:mask_manner}.
Note that performing $\texttt{argmin}(\cdot)$ operation over predicted reconstruction loss $\hat{\mathcal{L}}^t$ means we have generated a task even easier than the random baseline.
$\alpha_0 < \alpha_T$ indicates an easy-to-hard mask generation introduced in \cref{sec:mask}, while $\alpha_0 > \alpha_T$ means the opposite, \textit{i.e.}, a hard-to-easy manner, which is also studied in \cref{tab:mask_manner}.
All results verify the necessity of a hard pretext task and the easy-to-hard manner.
Both \texttt{argmin}$(\cdot)$ operation and the hard-to-easy mask generation manner leads to performance degradation over random masking baseline.

\paragraph{Predicting loss formulations.}
We study different designs of predicting loss in the following table, including absolute loss based on MSE introduced in \cref{eq:mse} and relative loss based on BCE defined in \cref{eq:bce}.
As expected, BCE is a better choice for mining \textit{relative relationship} between patches, instead of absolute values of reconstruction losses as MSE does, outperforming absolute MSE by +0.18\%.

\paragraph{Downstream tasks.}
We evaluate transfer learning performance using the pre-trained models in \cref{tab:target}, including COCO~\cite{lin2014microsoft} object detection and instance segmentation, and ADE20k~\cite{zhou2017scene} semantic segmentation.

As illustrated in \cref{tab:downstream}, equipped with our proposed \method, it outperforms +1.58 AP$_{\text{box}}$ and +1.14 AP$_{\text{mask}}$ on COCO~\cite{lin2014microsoft}, and +1.60 mIoU on ADE20k~\cite{zhou2017scene}, over MAE~\cite{he2022masked} baseline, \textit{i.e.}, taking raw RGB pixel as the learning target.
When using CLIP~\cite{radford2021learning} features as the learning target, it outperforms +0.36 AP$_{\text{box}}$ and +0.41 AP$_{\text{mask}}$ on COCO~\cite{lin2014microsoft}, and +0.76 mIoU on ADE20k~\cite{zhou2017scene} over baseline, respectively.

Notably, \textit{only} taking the predicting loss $\mathcal{L}_{\mathrm{pred}}$ as the extra objective manages to boost the performance across downstream tasks, verifying the effectiveness of making the model be the teacher, instead of only a student.
These observations are consistent across different learning targets.

\begin{table}[t]
    \centering
    \caption{
    Ablation study on different \textbf{mask strategies}.
    We study the effectiveness of the $\texttt{argmax}(\cdot)$ performed on predicted reconstruction loss $\hat{\mathcal{L}}^t$ and the ``easy-to-hard'' manner.
    Note that $\texttt{argmin}(\cdot)$ means that we mask those easy patches.
    %
    }
    \label{tab:mask_manner}
    \vspace{-8pt}
    \scalebox{0.85}{
    \begin{tabular}{l|l|lll|l}
    \toprule
    case & operation & $\gamma$ & $\alpha_0$ & $\alpha_T$ & fine-tune \\
    \midrule
    random & - & 75 & 0 & 0 & 82.49 \\
    \rowcolor{Light}
    learn to mask & \texttt{argmax}$(\cdot)$ & 75 & 0 & 0.5 & \textbf{82.95 \up{0.46}} \\
    learn to mask & \texttt{argmin}$(\cdot)$ & 75 & 0 & 0.5 & 82.36 \down{0.13} \\
    \midrule
    \midrule
    case & manner & $\gamma$ & $\alpha_0$ & $\alpha_T$ & fine-tune \\
    \midrule
    random & - & 75 & 0 & 0 & 82.49 \\
    \rowcolor{Light}
    learn to mask & easy-to-hard & 75 & 0 & 0.5 & \textbf{82.95 \up{0.46}} \\
    learn to mask & hard-to-easy & 75 & 0.5 & 0 & 81.71 \down{0.78} \\
    \bottomrule
    \end{tabular}}
\end{table}

\begin{table}[t]
    \centering
    \caption{
    Ablations on \textbf{predicting loss formulation}.
    We study the absolute loss introduced in \cref{eq:mse} and the relative loss described in \cref{eq:bce}.
    %
    }
    \vspace{-8pt}
    \setlength{\tabcolsep}{13pt}
    \scalebox{0.85}{
    \begin{tabular}{llll}
    \toprule
    case & fine-tune & \gc{linear} & \gc{$k$-NN} \\
    \midrule
    none (MAE~\cite{he2022masked}) & 82.23 & \gc{51.26} & \gc{31.98} \\
    absolute MSE & 82.77 \up{0.54} & \gc{51.85} & \gc{34.47} \\
    \rowcolor{Light}
    relative BCE & \textbf{82.95 \up{0.72}} & \textbf{\gc{54.92}} & \textbf{\gc{36.09}} \\
    \bottomrule
    \end{tabular}}
\end{table}

\begin{table}[t]
    \centering
    \caption{
    Ablations on \textbf{downstream tasks}.
    We take RGB and CLIP~\cite{radford2021learning} features as the learning target, representing \textit{pixel regression} and \textit{knowledge distillation} cases.
    All cases are first pre-trained 200 epochs on ImageNet-1K~\cite{russakovsky2015imagenet} with ViB-B/16~\cite{dosovitskiy2020image} followed by fine-tuning.
    %
    }
    \label{tab:downstream}
    \vspace{-8pt}
    \scalebox{0.83}{
    \setlength{\tabcolsep}{2.8pt}
    \begin{tabular}{l|ll|ll|l}
    \toprule
    \multirow{2}{*}{target} &
    \multirow{2}{*}{$\mathcal{L}_{\mathrm{pred}}$} &
    learn &
    \multicolumn{2}{l|}{COCO} & 
    \multicolumn{1}{l}{ADE20k} \\
    \cline{4-6}
    & & to mask & AP$_{\text{box}}$ & AP$_{\text{mask}}$ & mIoU \\
    \midrule
    \multirow{3}{*}{RGB} & 
    - & -  & 40.45 & 37.01 & 40.49 \\
    & \checkmark & - & 40.98 \up{0.53} & 37.34 \up{0.33} & 41.45 \up{0.96} \\
    & \cellcolor{Light}{\checkmark} & \cellcolor{Light}{\checkmark} & \cellcolor{Light}{\textbf{42.03 \up{1.58}}} & \cellcolor{Light}{\textbf{38.15 \up{1.14}}} & \cellcolor{Light}{\textbf{42.09 \up{1.60}}} \\
    \midrule
    \multirow{3}{*}{CLIP~\cite{radford2021learning}} & - & -  & 46.21 & 41.55 & 46.59 \\
    & \checkmark & - & 46.43 \up{0.22} & 41.80 \up{0.25} & 46.97 \up{0.38} \\
    & \checkmark & \checkmark & \textbf{46.57 \up{0.36}} & \textbf{41.96 \up{0.41}} & \textbf{47.35 \up{0.76}} \\
    \bottomrule
    \end{tabular}}
    \vspace{-10pt}
\end{table}

\subsection{Comparison with Previous Alternatives}
\label{sec:sota}

We compare our proposed \method with the supervised baseline and a wide range of self-supervised alternatives using fine-tuning accuracy in \cref{tab:sota},
where selected methods can be summarized into three mainstream: (1) contrastive learning methods~\cite{chen2021empirical, caron2021emerging}, (2) MIM with pixel regression methods~\cite{he2022masked, xie2022simmim}, and (3) MIM with feature distillation methods~\cite{zhou2021ibot, dong2022bootstrapped, bao2021beit}.
Effective pre-training epoch%
\footnote{Effective pre-training epochs accounts the actual trained images/views defined by~\cite{zhou2021ibot}.
Details can be found in \supp.} is used for fair comparison following~\cite{zhou2021ibot}.
%
%
All methods are evaluated under the same input size \textit{i.e.}, 224$\times$224.
We take raw RGB as the learning target following~\cite{he2022masked, xie2022simmim}.

Notably, with only 200 epochs pre-training, our \method achieves 83.0\% and 84.5\% Top-1 accuracy with ViT-B and ViT-L backbone, respectively, 
surpassing MAE~\cite{he2022masked} by +0.8\% and +1.2\%, and the supervised baseline by +2.1\% and +1.9\%, respectively.
With a longer training schedule, \textit{i.e.}, 800 epochs, \method achieves 84.2\% and 85.8\% Top-1 accuracy with ViT-B and ViT-L backbone, outperforming MAE~\cite{he2022masked} by +0.6\% and +0.7\%, respectively.
Strikingly, \method reaches comparable results with \textit{feature distillation} alternative BootMAE~\cite{dong2022bootstrapped}.
From \cref{tab:target}, taking EMA features as the learning target for \method, which is the same as BootMAE~\cite{dong2022bootstrapped}, can further improve the performance by $\sim$ 0.5\%.


\begin{table}[t]
    \centering
    \caption{
    Comparison with state-of-the-art alternatives on \textbf{ImageNet-1K}.
    All methods are evaluated by fine-tuning.
    The resolution of images is 224$\times$224 for both pre-training and fine-tuning.
    $\dag$ means our implementation.
    $\ddag$ means the result is borrowed from~\cite{he2022masked}.
    %
    }
    \label{tab:sota}
    \vspace{-8pt}
    \setlength{\tabcolsep}{10pt}
    \scalebox{0.85}{
    \begin{tabular}{ll lll}
    \toprule
    method & & eff. ep. & ViT-B & ViT-L \\
    \midrule
    \gc{scratch} & & \gc{-} & \gc{80.9$^{\dag}$} & \gc{82.6$^{\ddag}$} \\
    \midrule
    \multicolumn{5}{l}{\textit{Contrastive Learning}} \\
    MoCo v3$^{\ddag}$~\cite{chen2021empirical} & \pub{ICCV'21} & 600 & 83.2 & 84.1 \\
    DINO$^{\ddag}$~\cite{caron2021emerging} & \pub{ICCV'21} & 1600 & 83.6 & - \\
    \midrule
    \multicolumn{5}{l}{\textit{MIM with Pixel Regression}} \\
    MAE~\cite{he2022masked} & \pub{CVPR'22} & 200 & 82.2$^{\dag}$ & 83.3$^{\ddag}$ \\
    \rowcolor{Light}
    \method & \pub{Ours} & 200 & \textbf{83.0} & \textbf{84.5} \\
    MAE$^{\ddag}$~\cite{he2022masked} & \pub{CVPR'22} & 1600 & 83.6 & 85.1 \\
    SimMIM~\cite{xie2022simmim} & \pub{CVPR'22} & 800 & 83.8 & - \\
    \rowcolor{Light}
    \method & \pub{Ours} & 800 & \textbf{84.2} & \textbf{85.8} \\
    \midrule
    \multicolumn{5}{l}{\textit{MIM with Feature Distillation}} \\
    BEiT$^{\ddag}$~\cite{bao2021beit} & \pub{ICLR'22} & 800& 83.2 & 85.2 \\
    iBOT~\cite{zhou2021ibot} & \pub{ICLR'22} & 1600 & 84.0 & - \\
    BootMAE~\cite{dong2022bootstrapped} & \pub{ECCV'22} & 800 & 84.2 & 85.9 \\
    \bottomrule
    \end{tabular}}
\end{table}

\begin{table}[t]
    \centering
    \caption{
    Comparison with state-of-the-art alternatives on \textbf{ADE20k semantic segmentation} using UperNet.
    We take mIoU as the metric.
    $\ddag$ means the result is borrowed from~\cite{he2022masked}.
    }
    \label{tab:sota_seg}
    \vspace{-8pt}
    \setlength{\tabcolsep}{10pt}
    \scalebox{0.85}{
    \begin{tabular}{ll|cc}
    \toprule
    method & & ViT-B & ViT-L \\
    \midrule
    supervised$^{\ddag}$ & & 47.4 & 49.9 \\
    MoCo v3$^{\ddag}$~\cite{chen2021empirical} & \pub{ICCV'21} & 47.3 & 49.1 \\
    BEiT$^{\ddag}$~\cite{bao2021beit} & \pub{ICLR'22} & 47.1 & 53.3 \\
    MAE$^{\ddag}$~\cite{he2022masked} & \pub{CVPR'22} & 48.1 & 53.6 \\
    SemMAE~\cite{li2022semmae} & \pub{NeurIPS'22} & 46.3 & - \\
    \rowcolor{Light}
    \method & \pub{Ours} & \textbf{48.5} & \textbf{54.6} \\
    \bottomrule
    \end{tabular}}
    \vspace{-10pt}
\end{table}

\paragraph{Semantic Segmentation.}
We experiment on ADE20k~\cite{zhou2017scene} using UperNet~\cite{xiao2018unified} for 160k iterations  in \cref{tab:sota_seg}.
%
%
From the table, we can tell that our \method significantly improves performance over supervised pre-training by +1.1 mIoU (48.5 \textit{v.s.} 47.4) with ViT-B and +4.7 mIoU (54.6 \textit{v.s.} 49.9) with ViT-L, respectively.
More importantly, our \method outperforms self-supervised alternatives under all settings.
For example, with ViT-L, \method surpasses MAE~\cite{he2022masked} by +1.0 (54.6 \textit{v.s.} 53.6) mIoU.



\begin{figure}[t]
    \centering
    \includegraphics[width=1\linewidth]{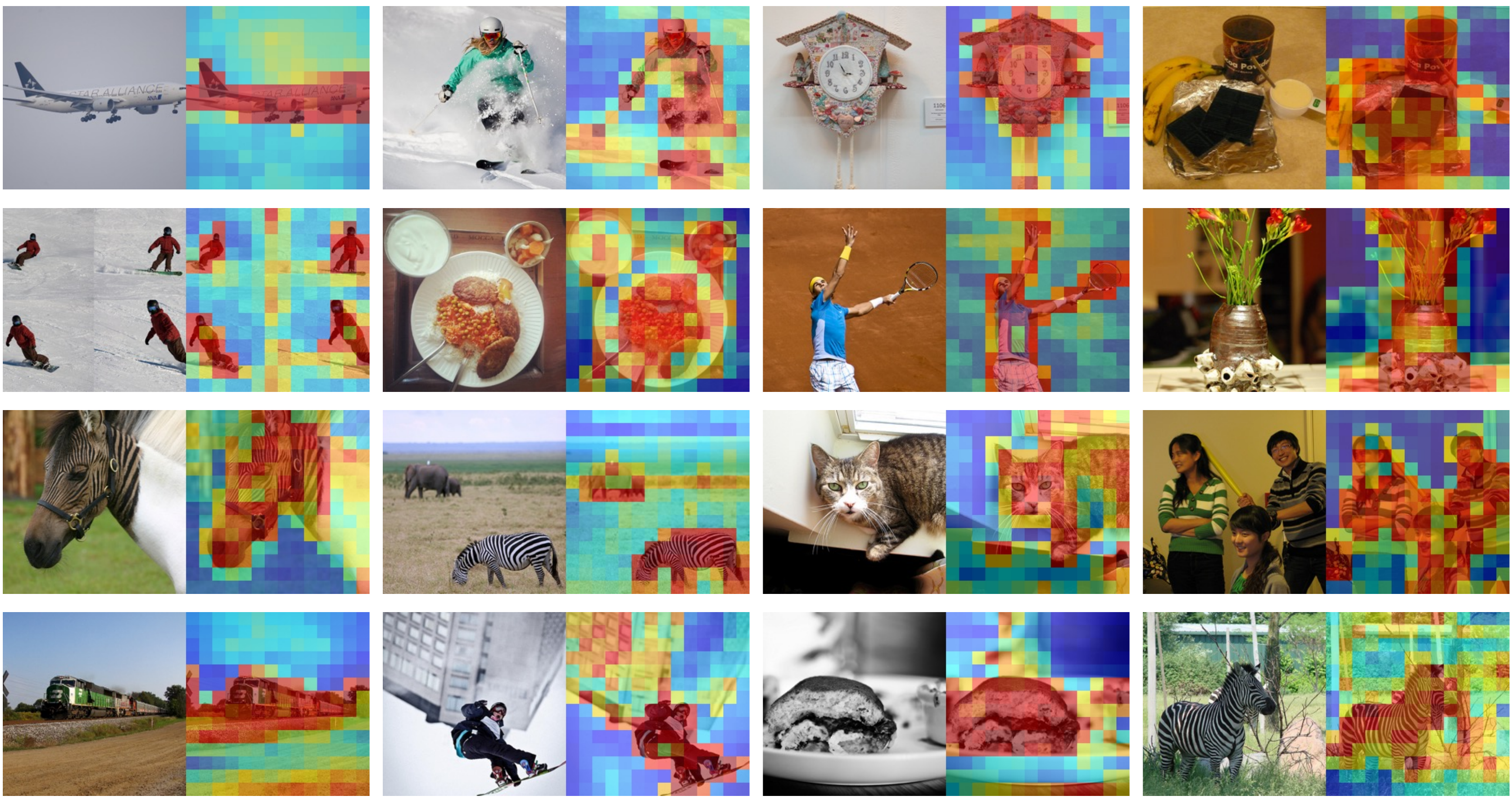}
    \vspace{-18pt}
    \caption{
    Visualization on \textbf{COCO} \textit{validation} set.
    For each tuple, we show the \textit{image} (left) and \textit{predicted} reconstruction losses (right).
    %
    }
    \label{fig:coco_half}
    \vspace{-10pt}
\end{figure}

\paragraph{Visualization of predicted losses.}
We provide qualitative results on COCO~\cite{lin2014microsoft} \textit{validation} set in \cref{fig:coco_half}, where the model has \textit{never} seen this dataset.
Patches with higher \textit{predicted} reconstruction loss usually are more discriminative.

\section{Conclusion}\label{sec:conclusion}

In this paper, we find it necessary to \textit{make the model stand in the shoes of a teacher} for MIM pre-training, and verify that the patch-wise reconstruction loss can naturally be the metric of the reconstruction difficulty.
To this end, we propose \method, which introduces an auxiliary reconstruction loss prediction task, and thus guides the training procedure iteratively in a produce-and-solve manner.
%
%
Experimentally, \method bootstraps the performance of masked image modeling across various downstream tasks.
Ablations across different learning targets show that \method, as a plug-and-play module, can be effortlessly incorporated into existing frameworks (\textit{e.g.}, pixel regression~\cite{he2022masked, xie2022simmim} and feature prediction~\cite{zhou2021ibot, dong2022bootstrapped, wei2022masked}) and bring \textit{consistent} performance improvements.
%

\paragraph{Broader impact.}
Techniques that mine hard examples are widely used in object detection~\cite{lin2017focal, shrivastava2016training, li2019gradient}.
Loss prediction can be a brand-new alternative.
Furthermore, it can be also used as a technique to filter high-quality pseudo-labels in label-efficient learning~\cite{wang2022semi, du2022learning, wang2023balancing}.
Meanwhile, as shown in \cref{fig:discriminative} and \cref{fig:coco_half}, \textit{the salient area tends to have a higher predicted loss}, and thus \method may also be used for saliency detection~\cite{wang2021salient} and unsupervised segmentation~\cite{van2022discovering, van2021unsupervised}.
We hope these perspectives will inspire future work.

\paragraph{Discussion.}
As a common problem of MIM, the performances of linear probing and $k$-NN classification are not as comparable as contrastive learning alternatives~\cite{he2022masked}.
%
%
In addition, \method needs more computation cost due to the extra decoder.
It takes $\sim$1.1$\times$ time to train our \method with ViT-L~\cite{dosovitskiy2020image} against MAE~\cite{he2022masked} baseline.
How to design a loss prediction task without an extra auxiliary decoder can be further studied.

\section*{Acknowledgements}

This work was supported in part by the Major Project for New Generation of AI (No.~2018AAA0100400), the National Natural Science Foundation of China (No.~61836014, No.~U21B2042, No.~62072457, No.~62006231), and the InnoHK program.

{\fontsize{8.2pt}{9.84pt}\selectfont
\bibliographystyle{ieee_fullname}
\bibliography{ref}
}
\clearpage
\renewcommand\thefigure{S\arabic{figure}}
\renewcommand\thetable{S\arabic{table}}  
\renewcommand\theequation{S\arabic{equation}}
\renewcommand\thealgorithm{S\arabic{algorithm}}
\setcounter{equation}{0}
\setcounter{table}{0}
\setcounter{figure}{0}
\setcounter{section}{0}
\setcounter{algorithm}{0}
\renewcommand\thesection{\Alph{section}}

\section*{Supplementary Material}

In this supplementary material, we first provide mode implementation details for reproducibility in \cref{sec:detail}.
Next, in \cref{sec:decoder}, we ablate baselines (\textit{i.e.}, BEiT~\cite{bao2021beit} and iBOT~\cite{zhou2021ibot}) and decoder designs.
The pseudo-code of the easy-to-hard mask generation in a Pytorch-like style is provided in \cref{sec:code}.
Finally, in \cref{sec:visual}, we provide both visual and quantitative evidence of our key assumption: \textit{discriminative patches are usually hard to reconstruct}.

\section{Implementation Details}
\label{sec:detail}

\noindent\textbf{ViT Architecture.}
We follow the standard vanilla ViT~\cite{dosovitskiy2020image} architecture used in MAE~\cite{he2022masked} as the backbone, which is a stack of Transformer blocks~\cite{vaswani2017attention}.
Following MAE~\cite{he2022masked} and UM-MAE~\cite{li2022uniform}, we use the sine-cosine positional embedding.
For the downstream classification task, we use features globally averaged from the encoder output for both end-to-end fine-tuning, linear probing, and $k$-NN classification.

\paragraph{Decoder Design.}
Our \method contains two decoders, \textit{i.e.}, the image reconstructor and the loss predictor.
These two decoders share the architecture, and each decoder is a stack of Transformer blocks~\cite{vaswani2017attention} followed by a linear projector.

\paragraph{Effective Training Epochs.}
Following iBOT~\cite{zhou2021ibot}, we take the effective training epochs as the metric of the training schedule, due to extra computation costs brought by multi-crop~\cite{caron2020unsupervised} augmentation,
which is a widely used technique for contrastive methods.
Specifically, the effective training epochs are defined as the actual pre-training
epochs multiplied with a scaling factor $r$.
For instance, DINO~\cite{caron2021emerging} is trained with 2 global 224$\times$224 crops and 10 local 96$\times$96 crops, and thus $r=2 + (96/224)^2 \times 10 \approx 4$.
More details and examples can be found in~\cite{zhou2021ibot}.

\subsection{ImageNet Classification}
For all experiments in this paper, we take ImageNet-1K~\cite{russakovsky2015imagenet}, which contains 1.3M images for 1K categories, as the pre-trained dataset.
By default, we take ViT-B/16~\cite{dosovitskiy2020image} as the backbone and it is pre-trained 200 epochs followed by 100 epochs of end-to-end fine-tuning.
Implementation details can be found in \cref{tab:pre}, \cref{tab:fine}, and \cref{tab:lin}.
Most of the configurations are borrowed from MAE~\cite{he2022masked}.
The linear learning rate scaling rule~\cite{goyal2017accurate} is adopted: $lr = lr_{\mathrm{base}} \times \mathrm{batch\_size}\ /\ 256$.
For supervised training from scratch, we simply follow the fine-tuning setting without another tuning.

We follow the linear probing setting of MoCo v3~\cite{chen2021empirical}.
We do not use mixup~\cite{zhang2017mixup}, cutmix~\cite{yun2019cutmix}, drop path~\cite{huang2016deep}, and color jitter.
The $k$-NN classification settings are borrowed from DINO~\cite{caron2021emerging}.
All images are first resized to 256$\times$256 and then center-cropped to 224$\times$224.
We report the best result among $k=10,20,100,200$.

\begin{table}[t]
    \centering
    \caption{
    \textbf{Pre-training settings}.
    By default, we use ViT-B/16~\cite{dosovitskiy2020image} as the backbone and apply 200 epochs pre-training.
    }
    \label{tab:pre}
    \vspace{-8pt}
    \setlength{\tabcolsep}{18pt}
    \scalebox{0.85}{
    \begin{tabular}{l|l}
    config & value \\
    \shline
    optimizer & AdamW~\cite{loshchilov2017decoupled} \\
    base learning rate & 1.5e-4 \\
    weight decay & 0.05 \\
    momentum & $\beta_1$, $\beta_2$ = 0.9, 0.95~\cite{chen2020generative} \\
    layer-wise lr decay~\cite{clark2020electra} & 1.0 \\
    batch size & 4096 \\
    learning rate schedule & cosine decay~\cite{loshchilov2016sgdr} \\
    warmup epochs & 10 (ViT-B), 40 (ViT-L) \\
    training epochs & 200 \\
    augmentation & RandomResizedCrop \\
    \end{tabular}}
\end{table}

\begin{table}[t]
    \centering
    \caption{
    \textbf{Fine-tuning settings}.
    By default, we use ViT-B/16~\cite{dosovitskiy2020image} as the backbone and apply 100 epochs fine-tuning on ImageNet-1K~\cite{russakovsky2015imagenet} after pre-training.
    }
    \label{tab:fine}
    \vspace{-8pt}
    \setlength{\tabcolsep}{10pt}
    \scalebox{0.85}{
    \begin{tabular}{l|l}
    config & value \\
    \shline
    optimizer & AdamW~\cite{loshchilov2017decoupled} \\
    base learning rate & 5e-4 \\
    weight decay & 0.05\\
    momentum & $\beta_1$, $\beta_2$ = 0.9, 0.999 \\
    layer-wise lr decay~\cite{clark2020electra} & 0.8 \\
    batch size & 1024 \\
    learning rate schedule & cosine decay~\cite{loshchilov2016sgdr} \\
    warmup epochs & 5 \\
    training epochs & 100 (ViT-B/16), 50 (ViT-L/16) \\
    augmentation & RandAug (9, 0.5)~\cite{cubuk2020randaugment} \\
    label smoothing~\cite{szegedy2016rethinking} & 0.1 \\
    mixup~\cite{zhang2017mixup} & 0.8 \\
    cutmix~\cite{yun2019cutmix} & 1.0 \\
    drop path~\cite{huang2016deep} & 0.1 \\
    \end{tabular}}
\end{table}

\begin{table}[t]
    \centering
    \caption{
    \textbf{Linear probing settings}.
    By default, we use ViT-B/16~\cite{dosovitskiy2020image} as the backbone and apply 100 epochs linear probing on ImageNet-1K~\cite{russakovsky2015imagenet} after pre-training.
    }
    \label{tab:lin}
    \vspace{-8pt}
    \setlength{\tabcolsep}{22pt}
    \scalebox{0.85}{
    \begin{tabular}{l|l}
    config & value \\
    \shline
    optimizer & SGD \\
    base learning rate & 1e-3 \\
    weight decay & 0 \\
    momentum & $\beta_1$ = 0.9 \\
    batch size & 4096 \\
    learning rate schedule & cosine decay~\cite{loshchilov2016sgdr} \\
    warmup epochs & 10 \\
    training epochs & 100 \\
    augmentation & RandomResizedCrop \\
    \end{tabular}}
\end{table}

\begin{table}[t]
    \centering
    \caption{
    Ablation study on different \textbf{decoder designs}.
    The speedup is evaluated under 8 Telsa V100 GPUs with 32 images with resolution 224$\times$224 per GPU.
    The default settings of our proposed \method are \colorbox{Light}{highlighted} in color.
    }
    \label{tab:decoder}
    \vspace{-8pt}
    \setlength{\tabcolsep}{12pt}
    \scalebox{0.85}{
    \begin{tabular}{c|llll}
    \toprule
    \# blocks & speedup & fine-tune & \gc{linear} & \gc{$k$-NN} \\
    \midrule
    1 & 1.94$\times$ & 82.67 & \gc{39.83} & \gc{16.83} \\
    2 & 1.68$\times$ & 82.50 & \gc{46.74} & \gc{22.63} \\
    4 & 1.37$\times$ & 82.75 & \gc{53.95} & \gc{33.60} \\
    \rowcolor{Light}
    8 & 1.00$\times$ & \textbf{82.95} & \gc{54.92} & \gc{36.09} \\
    12 & 0.76$\times$ & 82.84 & \gc{54.83} & \gc{35.93} \\
    \midrule
    \midrule
    \# dim & speedup & fine-tune & \gc{linear} & \gc{$k$-NN} \\
    \midrule
    128 & 1.31$\times$ & 82.74 & \gc{42.51} & \gc{17.67} \\
    256 & 1.18$\times$ & 82.80 & \gc{52.39} & \gc{29.46} \\
    \rowcolor{Light}
    512 & 1.00$\times$ & \textbf{82.95} & \gc{54.92} & \gc{36.09} \\
    1024 & 0.61$\times$ & 82.81 & \gc{54.01} & \gc{36.54} \\
    \bottomrule
    \end{tabular}}
    \vspace{-10pt}
\end{table}

\subsection{COCO Object Detection and Segmentation}

\noindent\textbf{Network Architecture.}
We take Mask R-CNN~\cite{he2017mask} with FPN~\cite{lin2017feature} as the object detector.
Following~\cite{he2022masked} and~\cite{li2022uniform}, to obtain pyramid feature maps for matching the requirements of FPN~\cite{lin2017feature}, whose feature maps are all with a stride of 16, we equally divide the backbone into 4 subsets, each consisting of a last global-window block and several local-window blocks otherwise, and then apply convolutions to get the intermediate feature maps at different scales (stride 4, 8, 16, or 32), which is the same as ResNet~\cite{he2016deep}.

\paragraph{Training.}
We perform end-to-end fine-tuning on COCO~\cite{lin2014microsoft} for 1$\times$ schedule, \textit{i.e.}, 12 epochs, for ablations (\textit{i.e.}, Tab. \textcolor{red}{6}) with 1024$\times$1024 resolution.
We simply follow the configuration of ViTDet~\cite{li2021benchmarking} in detectron2~\cite{wu2019detectron2}.
Experiments are conducted on 8 Telsa V100 GPUs with a batch size of 16.

\subsection{ADE20k Semantic Segmentation}

\noindent\textbf{Network Architecture.}
We take UperNet~\cite{xiao2018unified} as the segmentation decoder following the code of~\cite{bao2021beit, mmseg2020, li2022uniform}.

\paragraph{Training.}
Fine-tuning on ADE20k~\cite{zhou2017scene} for 80k iterations is performed for ablations.
When compared with previous methods, 160k iterations of fine-tuning are performed.
We adopt the exact same setting in mmsegmentation~\cite{mmseg2020}.
Specifically, each iteration consists of 16 images with 512$\times$512 resolution.
The AdamW~\cite{loshchilov2017decoupled} optimizer is adopted with an initial learning rate of 1e-4 and a weight decay of 0.05 with ViT-B.
For ViT-L, the learning rate is 2e-5.
We apply a polynomial learning rate schedule with the first warmup of 1500 iterations following common practice~\cite{li2022uniform, mmseg2020, bao2021beit}.
Experiments are conducted on 8 Telsa V100 GPUs.

\section{More Experiments}
\label{sec:decoder}

\begin{wraptable}{r}{0.37\linewidth}
    \centering
    \vspace{-13pt}
    \addtolength\leftskip{-15pt}
    \setlength{\tabcolsep}{3pt}
    \scalebox{0.85}{
    \begin{tabular}{ll}
    method & fine-tune \\
    \shline
    BEiT~\cite{bao2021beit} & 80.9 \\
    \method (w/ BEiT) & \textbf{81.5 \up{0.6}} \\
    iBOT~\cite{zhou2021ibot} & 82.9 \\
    \method (w/ iBOT) & \textbf{83.4 \up{0.5}} \\
    \end{tabular}}
    \vspace{-10pt}
\end{wraptable}

\noindent\textbf{\method over other baselines.}
We study the effectiveness of \method over BEiT~\cite{bao2021beit} and iBOT~\cite{zhou2021ibot} in the right table.
We perform 200 and 50 epochs pre-training for BEiT~\cite{bao2021beit} and iBOT~\cite{zhou2021ibot}, respectively.
Note that iBOT~\cite{zhou2021ibot} utilizes 2 global crops ($224^2$) and 10 local crops ($96^2$).
Therefore, the effective pre-training epoch of iBOT-based experiments is $50\times(2+\frac{10\times96^2}{224^2}) \approx 200$. 
From the table, we can tell that \method brings consistent improvements.

\paragraph{Ablations on decoder design.}
Our decoder is a stack of Transformer blocks~\cite{vaswani2017attention} with a fixed width following~\cite{he2022masked}.
We study its depth and width in \cref{tab:decoder}.
8 blocks with 512-d features is the best choice, which is exactly the same with MAE~\cite{he2022masked}.

\section{Implementation of Easy-to-Hard Masking}
\label{sec:code}

\cref{alg:mask} shows the implementation of easy-to-hard mask generation introduced in Sec. \textcolor{red}{3.4}.
Specifically, at training epoch $t$, we want to generate a binary mask $\mathbf{M}$ with $\gamma N$ patches to be masked.
Under the easy-to-hard manner, there are $\alpha_t \gamma N$ patches masked by predicted loss $\hat{\mathcal{L}}^t$ and the remaining $(1 - \alpha_t) \gamma N$ are randomly selected.

\begin{algorithm}[t]
\caption{Pseudo-Code of Easy-to-Hard Masking.}
\label{alg:mask}
\begin{lstlisting}[language=python]
# pred_t: predicted reconstruction loss
# t: current epoch
# T: total training epochs

# easy-to-hard mask generation
def mask_generation(pred_t, t, T, mask_ratio):
    L = len(pred_t)
    # total number of visible patches
    len_keep = int(L * (1 - mask_ratio))
    
    # number of patches masked by predicted loss
    alpha_t = alpha_0 + t/T * (alpha_T - alpha_0)
    len_pred = int(L * mask_ratio * alpha_t)
    ids_shuffle = argsort(pred_t)
    
    # compute remaining patches
    remain = delete(arange(L) - ids_shuffle[-len_pred:])
    
    # random masking for remained patches
    ids_shuffle[:(L-len_pred)] = shuffle(remain)
    
    # generate mask: 0 is remove, 1 is keep
    mask = ones([L,]).bool()
    mask[:len_keep] = 1
    
    # restore the mask
    ids_restore = argsort(ids_shuffle)
    return gather(mask, ids_restore)
\end{lstlisting}
\end{algorithm}

\section{Hard to Reconstruct \textit{v.s.} Discrimination}
\label{sec:visual}

\begin{figure*}[t]
    \centering
    \includegraphics[width=1\linewidth]{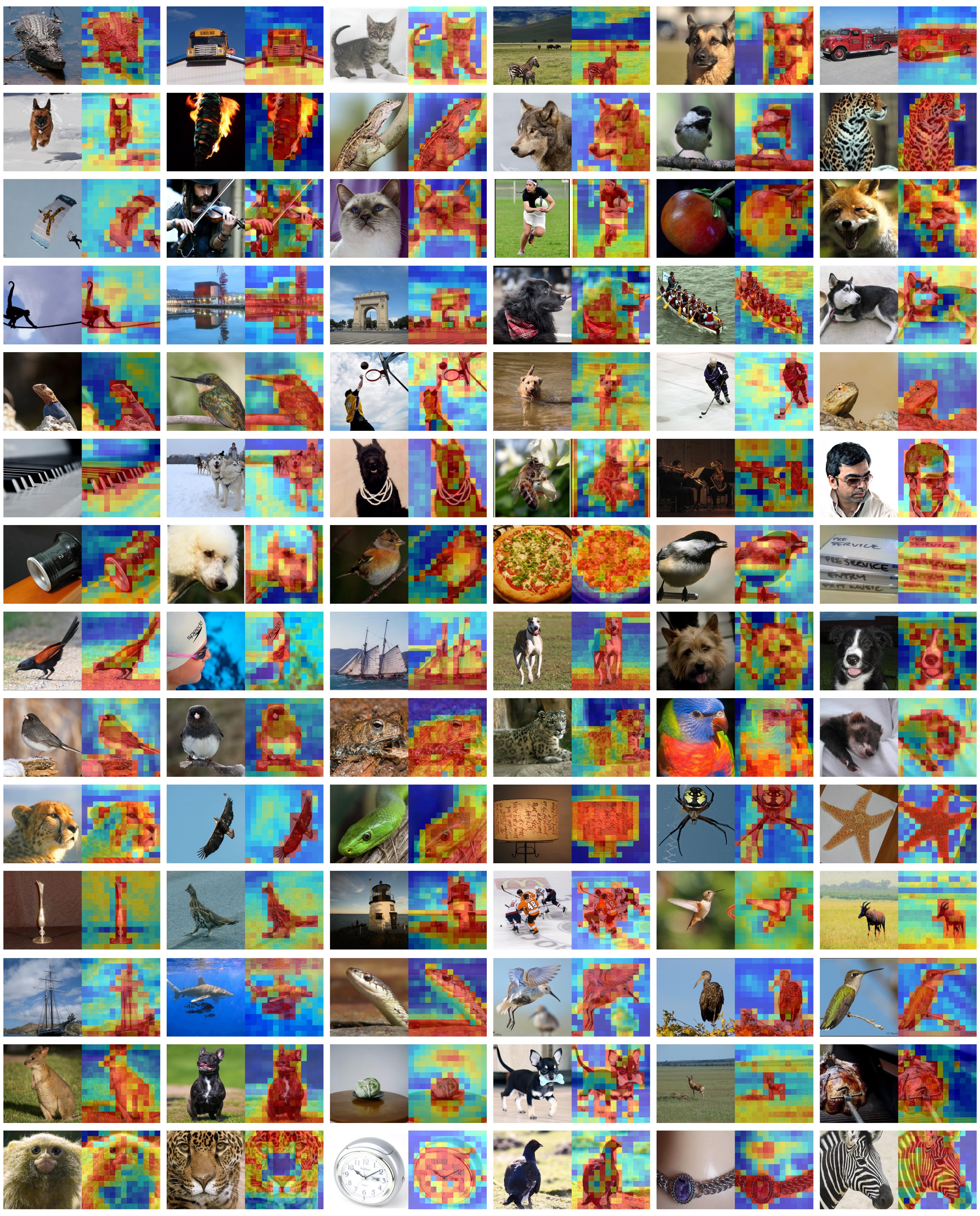}
    \vspace{-18pt}
    \caption{
    Qualitative results on \textbf{ImageNet-1K} \textit{validation} set.
    For each tuple, we show the \textit{input image} (left) and the patch-wise \textit{predicted} reconstruction loss (right).
    Red means higher losses and blue indicates the opposite.
    }
    \label{fig:imagenet}
\end{figure*}

\begin{figure*}[t]
    \centering
    \includegraphics[width=1\linewidth]{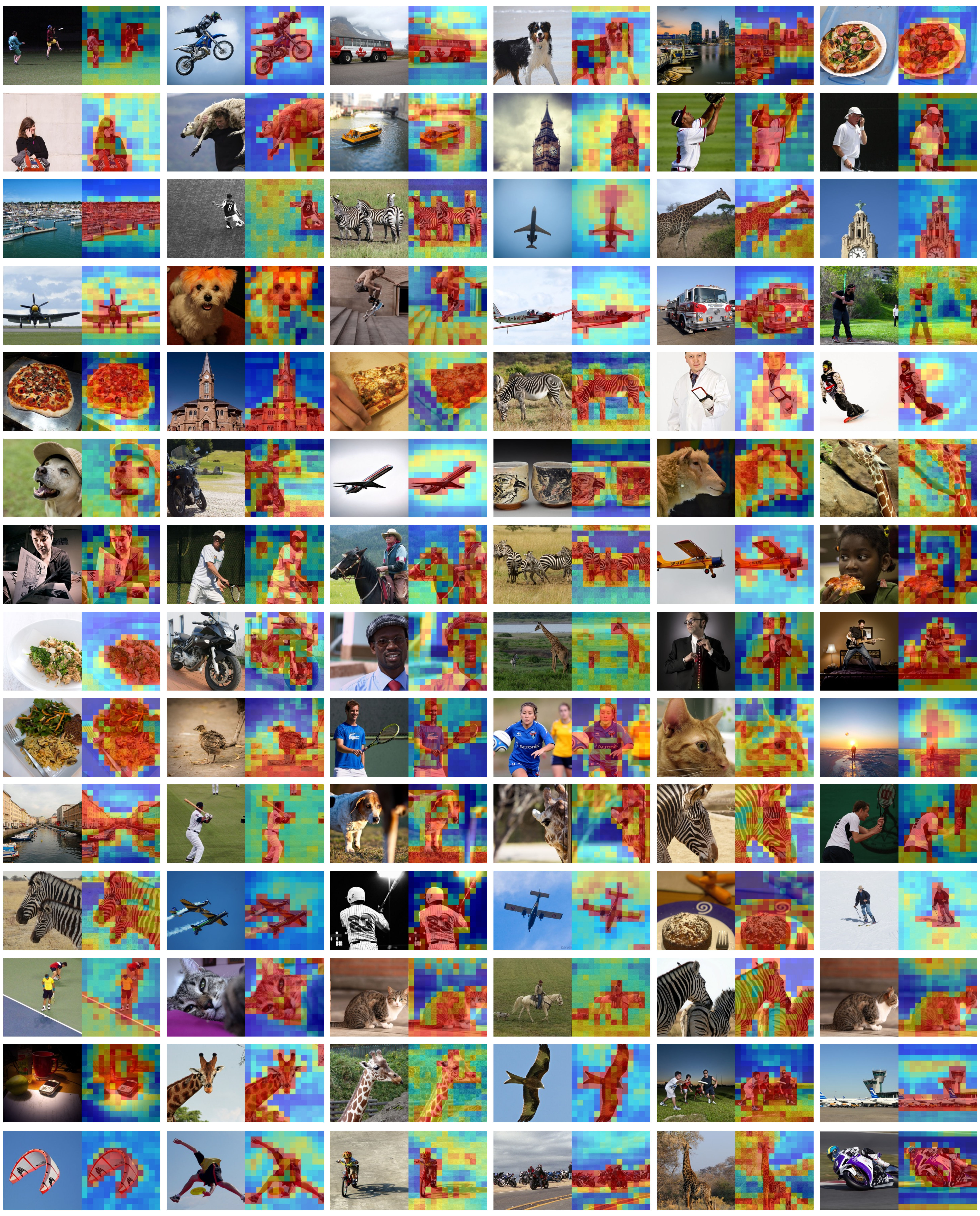}
    \vspace{-18pt}
    \caption{
    Qualitative results on \textbf{COCO} \textit{validation} set.
    For each tuple, we show the \textit{input image} (left) and the patch-wise \textit{predicted} reconstruction loss (right).
    Red means higher losses and blue indicates the opposite.
    }
    \label{fig:coco}
\end{figure*}

\noindent\textbf{Visual evidence.}
We provide qualitative results on ImageNet-1K~\cite{russakovsky2015imagenet} \textit{validation} set in \cref{fig:imagenet} and COCO~\cite{lin2014microsoft} \textit{validation} set in \cref{fig:coco}, respectively.
As illustrated in \cref{fig:coco,fig:imagenet}, patches with higher \textit{predicted} reconstruction loss usually are more discriminative (\textit{i.e.}, object or forehead).

\begin{wraptable}{r}{0.32\linewidth}
    \centering
    \vspace{-13pt}
    \addtolength\leftskip{-15pt}
    \setlength{\tabcolsep}{3pt}
    \scalebox{0.85}{
    \begin{tabular}{ll}
    input & accuracy \\
    \shline
    random 50\% & 79.1 \\
    bottom 50\% & 78.7 \down{0.4} \\
    top 50\% & \textbf{79.8 \up{0.7}} \\
    \gc{all 100\%} & \gc{80.9} \\
    \end{tabular}}
    \vspace{-10pt}
\end{wraptable}

\paragraph{Quantitative evidence.}
Here, we present a toy experiment to explore the relationship between \textit{hard to reconstruct} and \textit{discrimination for classification}.
In the right table, three ViT-B/16~\cite{dosovitskiy2020image} models are trained from scratch on ImageNet-1K for 100 epochs under image-level supervision.
\textit{Only 50\% patches are input}, and ``bottom'' and ``top'' indicates patches with lower and higher $\mathcal{L}_{\mathrm{pred}}$ are visible, respectively.
We load HPM pre-trained with 200 epochs for computing $\mathcal{L}_{\mathrm{pred}}$.
Empirically, patches with higher $\mathcal{L}_{\mathrm{pred}}$ contribute more to classification.
We hope this will inspire future work.

\end{document}